\documentclass[letterpaper, 10 pt, conference]{ieeeconf}  

\usepackage[utf8]{inputenc}
\usepackage{textcomp}
\usepackage{graphicx}
\usepackage[version=4]{mhchem}
\usepackage{siunitx}
\usepackage{longtable,tabularx}
\usepackage{accents}
\usepackage{comment}
\usepackage{mathtools}
\usepackage{subfigure}
\usepackage{caption}
\usepackage{color}
\usepackage{amsmath,amssymb}
\usepackage{booktabs}
\usepackage{tabu}
\usepackage{algorithm}
\usepackage{algpseudocode}
\usepackage{hyperref}
\usepackage[dvipsnames]{xcolor}
\usepackage{multirow}
\usepackage{enumerate}
\usepackage{svg}
\usepackage{epstopdf}
\usepackage{soul}
\usepackage{arydshln}
\usepackage[colorinlistoftodos]{todonotes}
\usepackage{colortbl}
\usepackage{import}

\usepackage{tikz}
\usepackage{siunitx} 
\usetikzlibrary{arrows, arrows.meta}
\usetikzlibrary{graphs}
\usetikzlibrary{snakes,backgrounds,patterns,matrix,shapes,fit,calc,shadows,plotmarks} 
\usetikzlibrary{bayesnet}
\usetikzlibrary{backgrounds}

\usepackage{standalone}

\tikzstyle{target}=[draw,fill=yellow!50,circle,minimum size=16pt,inner sep=0pt]

\tikzstyle{output}=[draw,fill=blue!50,circle,minimum size=16pt,inner sep=0pt]
\tikzstyle{bias}=[draw,fill=gray!50,circle,minimum size=20pt,inner sep=2pt]
\tikzstyle{arrow}=[arrows={{Latex[scale=0.5]}-}, thick]  
\tikzstyle{box}=[rectangle, draw=black!100] 

\tikzset{
    between/.style args={#1 and #2}{
         at = ($(#1)!0.5!(#2)$)
    }
}

\newcommand{\etal}{\emph{et al. }}

\IEEEoverridecommandlockouts                              

\title{\LARGE \bf
Towards Decentralized Heterogeneous Multi-Robot SLAM and Target Tracking
}

\author{Ofer Dagan$^{1}$, Tycho L. Cinquini$^{1}$, Luke Morrissey$^{2}$, Kristen Such$^{2}$, Nisar R. Ahmed$^{1}$, Christoffer Heckman$^{2}$
\thanks{
$^{1}$ Ofer Dagan, Tycho L. Cinquini, and Nisar R. Ahmed are with the Smead Aerospace Engineering Sciences Department, University of Colorado Boulder, Boulder, CO 80309 USA {\tt\small ofer.dagan@colorado.edu; nisar.ahmed@colorado.edu}
$^{2}$ Luke Morrissey, Kristen Such, and Christoffer Heckman are with the Computer Science Department, University of Colorado Boulder, Boulder, CO 80309 USA {\tt\small kristen.such@colorado.edu; christoffer.heckman@colorado.edu}}
}

\begin{document}

\maketitle
\thispagestyle{empty}
\pagestyle{empty}

\section{INTRODUCTION}

In many robotics problems, there is a significant gain in collaborative information sharing between multiple robots, for exploration, search and rescue, tracking multiple targets, or mapping large environments.  
In particular, there are different approaches to solving multi-robot state estimation problems:
they can be solved in a centralized manner, where all robots send their data to a fusion center that serves all robots; or they can be solved in a distributed or decentralized manner using algorithms such as consensus \cite{olfati-saber_consensus_2004} or decentralized data fusion (DDF) \cite{chong_distributed_1983}.
To the best of our knowledge, all of these approaches have a common assumption -- that each robot implements the same (homogeneous) underlying estimation algorithm. 
However, in practice, we want to allow collaboration between robots possessing different capabilities and that therefore must rely on heterogeneous algorithms.

We present a system architecture and the supporting theory to enable collaboration in a decentralized network of robots, where each robot relies on different estimation algorithms.
To develop our approach, we focus on the application to multi-robot simultaneous localization and mapping (SLAM) with multi-target tracking.
Our theoretical framework builds on our idea of exploiting the conditional independence structure inherent to many robotics applications \cite{dagan_exact_2023} to separate between each robot's local inference (estimation) tasks and fuse only relevant parts of their non-equal, but overlapping probability density function (pdfs). 
We rely on the widely used factor framework \cite{frey_factor_1997}, and its advantages in representing conditional independence to develop easily scalable, statistically consistent heterogeneous fusion algorithms \cite{dagan_factor_2021,dagan_conservative_2022}.
These algorithms significantly reduce local communication and computation requirements and enable robots to share information for SLAM with dense and metric-semantic maps while tracking dynamic targets and using different sensor suites. 
The ideas and work presented here demonstrate progress 
to allow a network of robots to collaboratively share information, irrespective of their local SLAM and tracking algorithms.



In multi-robot SLAM, most works are based on sparse landmark SLAM where robots share and fuse sub-maps e.g., \cite{nettleton_decentralised_2006}, \cite{reece_robust_2005}, \cite{julier_using_2007}, \cite{cunningham_ddf-sam_2013}.
In \cite{nettleton_decentralised_2006}, Nettleton \etal selects informative parts of the map to communicate between robots. 
They use a hybrid channel filter (CF) -- covariance intersection (CI) algorithm to ensure that previously communicated map data is counted not more than once. 
Reece and Roberts \cite{reece_robust_2005} improve upon this work by replacing the CI with a less conservative fusion rule called Bounded Covariance Inflation. 
In \cite{julier_using_2007} Julier and Uhlmann use different variations of the CI algorithm \cite{julier_non-divergent_1997} to solve the Full Covariance SLAM problem. 
Their approach suggests building three types of maps: main, relative, and local, using different instantiations of a Kalman filter (KF) and then fusing them together using CI to account for unknown dependencies between the estimates.    
In \cite{tchuiev_distributed_2020}, Tchiev and Indelman consider the problem of semantic distributed SLAM with sparse landmarks, where robots fuse belief over both the map and classes of objects. 
To avoid double counting previously communicated data and simplify bookkeeping, they assume the communication network topology is undirected and acyclic. 

While for sparse landmark SLAM it is possible to share parts of the map, i.e. coordinates of landmarks, this approach is less feasible when considering dense maps \cite{shan_lio-sam_2020}, \cite{campos_orb-slam3_2021}  and metric-semantic SLAM algorithms \cite{rosinol_kimera_2021}.
For that reason, recent approaches either use a centralized server to merge maps \cite{schmuck_ccm-slam_2019}, \cite{van_opdenbosch_collaborative_2019}, or choose to  collaboratively optimize the robots' trajectories (instead of the map) using pose graph optimization (PGO) techniques \cite{tian_kimera-multi_2022}. The improved trajectory estimate is then fed back to correct the local map estimate. 

The problem of collaborative tracking and SLAM has gotten much less attention in the literature than collaborative SLAM.  
In \cite{wang_online_2003} Wang \etal formulates the SLAM and detection and tracking of moving objects (DATMO) problem in Bayesian terms. 
For a single robot problem, they suggest splitting the SLAM and DATMO problems into two separate solvers but assume that data regarding the moving object (target) does not carry any information concerning the map and robot pose.
Moratuwage \etal \cite{moratuwage_collaborative_2013} use random finite sets (RFS) for collaborative multi-vehicle SLAM with moving object tracking. 
Their work solves for sparse landmark representation of the map and is not fully decentralized, as the posterior pdf is based on all measurements and trajectories.

In this paper, we present a new decentralized graph-based approach to the multi-robot SLAM and tracking problem.
We leverage factor graphs to split between different parts of the problem for efficient data sharing between robots in the network while enabling robots to use different local sparse landmark/dense/metric-semantic SLAM algorithms.
The paper is organized as follows: Sec. \ref{sec:probStatement} defines the decentralized heterogeneous multi-robot SLAM and tracking problem, Sec. \ref{sec:techApproach} presents our factor graph-based technical approach, and Sec. \ref{sec:sim} describes our planned simulation study and a roadmap for future hardware experiments.


\section{Problem Statement}
\label{sec:probStatement}
Consider a network of $n_r=|N_r|$ robots in an unknown joint environment. 
Each robot reasons about the environment and tries to gain situational awareness using its local, potentially different, SLAM algorithm.  
In addition, the robots are tasked with jointly inferring (estimating) the unknown state of a subset of $n_t=|N_t|$ static or dynamic targets at time step $k$.
In a probabilistic approach, the uncertainty over full system multi-robot SLAM and multi-target tracking problem can be described by,
\begin{equation}
     p(\chi|Z_k)=p(X_{k:0},M,T_k|Z_k).
     \label{eq:fullSLAM}
\end{equation}
Here, $p(\chi|Z_k)$ is the pdf over the full system set of random variables (rvs), conditioned on the data set $Z_k=\bigcup_{i\in N_r} Z^i_k$ gathered by all robots. 
$X_{k:0}$ is the set of rvs describing the uncertain poses of all $n_r$ robots at time steps $0$ to $k$, $T_k$ is the set of rvs describing the uncertain position of all $n_t$ targets at time step $k$, and $M$ is a set of rvs representing the map. 
Note that we defined $M$ rather ambiguously, so as to enable heterogeneity in the system. This allows each robot to hold a different representation of the environment, e.g. sparse landmarks, point clouds, and labeled objects.

\subsection*{Decentralized SLAM and Tracking:}
In a decentralized formulation of the problem, each robot $i$ is tasked with inferring a subset of the full set of rvs $\chi^i \subseteq \chi$: its own pose $x^i_{k:0}\subset X_{k:0}$, local map $M^i\subseteq M$, and the states $T^i$ of a subset $N_t^i\subseteq N_t$ of the targets. 
The local inference task of any robot $i\in N_r$ can be described by,
\begin{equation}
    p^i(\chi^i|Z^i_k)=p(x^i_{k:0},M^i,T^i_{k}|Z^i_k).
    \label{eq:decSLAM}
\end{equation}
In decentralized SLAM and tracking, each robot maintains a pdf (\ref{eq:decSLAM}) based on the data it gathers from local sensors and the data it receives from neighboring robots through peer-to-peer communication. 
Since each robot maintains a different, but overlapping pdf with its neighbors, this is an instance of a heterogeneous fusion problem \cite{dagan_exact_2023}.
In heterogeneous fusion, we split the robot's set of rvs $\chi^i$ to 
common rvs $\chi^i_C=\bigcup_{j\in N_r^i}^{}\chi^{ij}_C$, local rvs $\chi^i_L$ which are not monitored by any other robot in the network, and non-mutual rvs $\chi^{i\setminus j}=\chi^i_L\cup\{\chi^i_C\setminus \chi^{ij}_C \}$. 
In our proposed SLAM and tracking, the common rvs are the common target states $T^{ij}_C$, the local rvs are the local map, and the robot's pose states; and $T^{i\setminus j}$ is the subset of targets monitored by robot $i$, and possibly other robots, but not $j$.


\section{Technical Approach}
\label{sec:techApproach}

\begin{figure}
    \centering
    \includegraphics{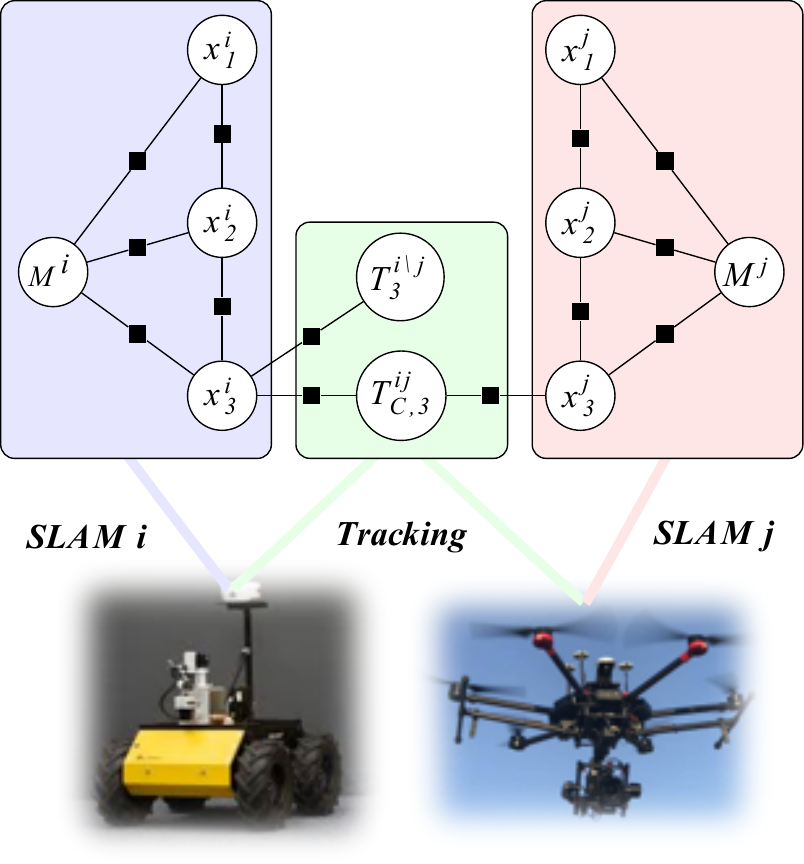}
    \caption{Factor graph representation of a 2-robot SLAM and target tracking application. Both robots $i$ and $j$ are building local maps $M^i$ and $M^j$ while cooperatively tracking targets in common $T^{ij}_C$ (observed at time step 3). The data  fused regarding common targets will indirectly update the map and the estimate of non-mutual targets $T^{i\setminus j}$.}
    \label{fig:approachGraph}
\end{figure}

The key part of our approach is exploiting the probabilistic conditional independence structure of the problem to: (i) enable robots to use different SLAM algorithms and only share `relevant' parts of their pdfs with each other for fusion, and (ii) enable inner-loop robot architecture that separates the SLAM solution, which includes the local map and ego pose, from the target tracking solution.
First, let us look at the example factor graph in Fig. \ref{fig:approachGraph} showing the dependency structure of a two-robot problem.

A factor graph \cite{frey_factor_1997} is an undirected bipartite graph $\mathcal{F}=(U,V,E)$ that represents a function, proportional to the joint pdf over all random variable nodes $v_m\in V$, and factorized into smaller functions given by the factor nodes $f_l\in U$. 
An edge $e_{lm}\in E$ in the graph only connects a factor node \emph{l} to a variable node \emph{m}.
The joint distribution over the graph is then proportional to the global function $f(V)$:
\begin{equation}
    p(V)\propto f(V)=\prod_{l}f_l(V_l),
    \label{eq:factorization}
\end{equation}
where $f_l(V_l)$ is a function of only those variables $v_m\in V_l$ connected to the factor \emph{l}. 
That enables the graph to explicitly express the conditional independence structure of the problem and makes factor graphs an attractive representation for decentralized inference problems \cite{dagan_factor_2021}.

The scenario depicted in the graph (Fig. \ref{fig:approachGraph}) shows two robots $i$ and $j$, tasked with a SLAM problem -- estimating their pose $x^i_{k:1} \ (x^j_{k:1})$ and the local map $M^i \ (M^j)$. 
At the same time, the two robots are tasked with estimating the state of a set of common targets $T_C^{ij}$ and a set of targets, only tracked by robot $i$, $T^{i\setminus j}$. 
From the graph, we see that non-mutual variables (pose, map, exclusive targets) of each robot are conditionally independent given the common target variables $T^{ij}_{C}$. 
With this conditional independence structure, we can use the heterogeneous fusion rule developed in \cite{dagan_exact_2023}, to perform a peer-to-peer fusion of data regarding common targets,
\begin{equation}
    \begin{split}
        &p_f^i(\chi^i|Z^{i,+}_k)\propto \\
        &p^i(x^i_{k:0}, M^i,T^{i\backslash j}|T^{ij}_{C,k},Z^{i,-}_k)\cdot
        \frac{p^i(T^{ij}_{C,k}|Z^{i,-}_k)p^j(T^{ij}_{C,k}|Z^{j,-}_k)}{p^{ij}_c(T^{ij}_{C,k}|Z^{i,-}_k \cap Z^{j,-}_k)}.  
    \end{split}
    \label{eq:Heterogeneous_fusion}
\end{equation}
Here $Z^{i,-}_k$ and $Z^{i,+}_k$ are the data sets available for agent $i$ at time $k$ prior and post fusion, respectively. 

\begin{figure*}[tb]
\centering
\resizebox{7.2in}{2.5in}{
\begin{tikzpicture}[new set=import nodes]
 \begin{scope}[nodes={set=import nodes}]
     \node (a) at (-1.3,3.5) {$(a)$};
     \node (x3t)[latent] at (2.5,1) {$x_3^i$};
     
     \node [latent, above=of x3t] (T3) {$t_3^t$};
     \node [factor, fill=red!100, between=x3t and T3] (ft3) {};
     \node [latent, left=of x3t, yshift=0.6cm, xshift=-1.35cm] (x3cf) {$x_3^i$};
     \node [factor, fill=blue!100, left=of x3t] (fs3) {};
     \node [factor, fill=blue!100, below=of x3cf, yshift=0.15cm] (fs3cf) {};

      \node (x1)[latent] at (-1,-1.0) {$x_1^i$};
      \node [factor, right=of x1] (f1) {};
      \node [latent, right=of f1, xshift=-0.55cm] (x2) {$x_2^i$};
      \node [factor, right=of x2] (f2) {};
      \node [latent, right=of f2, xshift=-0.55cm] (x3) {$x_3^i$};

     \node [latent, below=of x3,xshift=-0.5cm] (l1) {$M^i$};
     \node [factor,  between=x1 and l1] (fl1) {};
     \node [factor, between=x2 and l1] (fl2) {};
     \node [factor,  between=x3 and l1] (fl3) {};

     \begin{pgfonlayer}{background}
	\draw[fill=blue!10,rounded corners] (-1.5,-3.25) rectangle (3,0)
        node[below, xshift=-3cm, yshift=-2.75cm] {SLAM Module};
        \draw[fill=green!10,rounded corners] (-1.5,3.25) rectangle (3,0.5)      node[above, xshift=-3cm, yshift=2.2cm] {Tracking Module};
        
        \draw[fill=red!10,rounded corners] (0.15,2.5) rectangle (-1.35,0.65) node[above, xshift=0.35cm, yshift=1.35cm] {CF};
	
     \end{pgfonlayer}

  \end{scope}
  \graph {
    (import nodes);
    x3t--ft3, ft3--T3, 
    x1--f1, f1--x2, x2--f2, f2--x3,
    {x1,l1}--fl1, {x2,l1}--fl2,{x3,l1}--fl3,
    x3t--fs3, x3cf--fs3cf,
    fs3<-[dashed]x3, fs3cf<-[dashed]x3,

   };
   \begin{scope}[nodes={set=import nodes}]
     \node (b) at (3.4,3.5) {$(b)$};
     \node (x3t)[latent] at (7.2,1) {$x_3^i$};
     
     \node [latent, above=of x3t] (T3) {$t_3^t$};
     \node [factor, fill=red!100, between=x3t and T3] (ft3) {};
     \node [latent, left=of x3t, yshift=0.6cm, xshift=-1.35cm] (x3cf) {$x_3^i$};
     \node [factor, fill=blue!100, left=of x3t] (fs3) {};
     \node [factor, fill=blue!100, below=of x3cf, yshift=0.15cm] (fs3cf) {};
     \node [factor, fill=orange!100, right=of x3cf] (f3cf) {};
     \node  (e1)[const, left=of ft3, xshift=0.75cm]  {};

      \node (x1)[latent] at (3.7,-1.0) {$x_1^i$};
      \node [factor, right=of x1] (f1) {};
      \node [latent, right=of f1, xshift=-0.55cm] (x2) {$x_2^i$};
      \node [factor, right=of x2] (f2) {};
      \node [latent, right=of f2, xshift=-0.55cm] (x3) {$x_3^i$};
      \node [factor, fill=orange!100, above=of x3] (f3) {};

     \node [latent, below=of x3,xshift=-0.5cm] (l1) {$M^i$};
     \node [factor,  between=x1 and l1] (fl1) {};
     \node [factor, between=x2 and l1] (fl2) {};
     \node [factor,  between=x3 and l1] (fl3) {};
     

     \begin{pgfonlayer}{background}
	\draw[fill=blue!10,rounded corners] (3.2,-3.25) rectangle (7.7,0)
        node[below, xshift=-3cm, yshift=-2.75cm] {SLAM Module};
        \draw[fill=green!10,rounded corners] (3.2,3.25) rectangle (7.7,0.5)      node[above, xshift=-3cm, yshift=2.2cm] {Tracking Module};
        
        \draw[fill=red!10,rounded corners] (5.35,2.5) rectangle (3.35,0.65) node[above, xshift=0.35cm, yshift=1.35cm] {CF};
	
     \end{pgfonlayer}

  \end{scope}
  
 \graph {
    (import nodes);
    x3t--ft3, ft3--T3, 
    x1--f1, f1--x2, x2--f2, f2--x3,
    {x1,l1}--fl1, {x2,l1}--fl2,{x3,l1}--fl3,
    x3t--fs3, x3cf--fs3cf, x3--f3,
    x3cf--f3cf,
    f3cf->[dashed]f3,
    e1->[dashed]f3cf,

   };

   \begin{scope}[nodes={set=import nodes}]
     \node (c) at (8.1,3.5) {$(c)$};
     \node (x3t)[latent] at (11.15,1) {$x_3^i$};
     
     \node [latent, right=of x3t] (x5t) {$x_5^i$};
     \node [latent, above=of x5t] (T5) {$t_5^t$};
     \node [factor, between=x3t and T5] (ft3) {};
     \node [factor, fill=red!100, between=x5t and T5] (ft5) {};
     \node [factor,  left=of T5] (f03) {};
     \node [factor, fill=blue!100, between=x3t and x5t] (fs35) {};

     \node [latent, left=of x3t, yshift=0.6cm, xshift=-0.9cm] (x3cf) {$x_3^i$};

     \node [latent, right=of x3cf, xshift=-0.25cm] (x5cf) {$x_5^i$};
     
     \node [factor, fill=blue!100, between=x3cf and x5cf] (fs35cf) {};
     \node [factor, fill=orange!100, below=of x3cf, yshift=0.2cm] (f3cf) {};

      \node (x3)[latent] at (9.4,-1.0) {$x_3^i$};
      \node [factor, right=of x3] (f1) {};
      \node [latent, right=of f1, xshift=-0.55cm] (x4) {$x_4^i$};
      \node [factor, right=of x4] (f2) {};
      \node [latent, right=of f2, xshift=-0.55cm] (x5) {$x_5^i$};
      \node [factor, fill=blue!60, between=x3 and x5, yshift=0.75cm] (fm35) {};
      \node [factor, fill=orange!100, above=of x3, yshift=-0.2cm] (f3t) {};

       \node [draw, fill=black!100,circle,minimum size=4pt, left=of x3,xshift=0.75cm,inner sep=0pt] (dot0) {};
      \node [draw, fill=black!100,circle,minimum size=4pt, left=of x3,xshift=0.5cm,inner sep=0pt] () {};
      \node [draw, fill=black!100,circle,minimum size=4pt, left=of x3,xshift=0.25cm,inner sep=0pt] (dot1) {};

     \node [latent, below=of x4,xshift=-0.0cm] (l1) {$M^i$};
     \node [factor,  between=x3 and l1] (fl1) {};
     \node [factor, between=x4 and l1] (fl2) {};
     \node [factor,  between=x5 and l1] (fl3) {};

     \begin{pgfonlayer}{background}
	\draw[fill=blue!10,rounded corners] (7.9,-3.25) rectangle (13.4,0)
        node[below, xshift=-4cm, yshift=-2.75cm] {SLAM Module};
        \draw[fill=green!10,rounded corners] (7.9,3.25) rectangle (13.4,0.5)  node[above, xshift=-4cm, yshift=2.2cm] {Tracking Module};
        
        \draw[fill=red!10,rounded corners] (10.45,2.5) rectangle (8.05,0.65) node[above, xshift=0.35cm, yshift=1.35cm] {CF};

     \end{pgfonlayer}

  \end{scope}
  
 \graph {
    (import nodes);
    x3--[dashed]dot0,
    x3t--ft3, ft3--T5, f03--T5, x3t--fs35, x5t--fs35,
    {x5t,T5}--ft5, x3--f3t,
    x3--f1, f1--x4, x4--f2, f2--x5,
    {x3,l1}--fl1, {x4,l1}--fl2,{x5,l1}--fl3,
    fs35cf<-[dashed]fm35, fs35<-[dashed]fm35,
    x3--[dashed, color=black!100, bend left=20]fm35,
    x5--[dashed,  color=black!100, bend right=20]fm35,
    {x3cf,x5cf}--fs35cf, f3cf--x3cf,

   };
   \begin{scope}[nodes={set=import nodes}]
     \node (d) at (13.8,3.5) {$(d)$};
     \node (x3t)[latent] at (16.85,1) {$x_3^i$};
     
     \node [latent, right=of x3t] (x5t) {$x_5^i$};
     \node [latent, above=of x5t] (T5) {$t_5^t$};
     \node [factor, between=x3t and T5] (ft3) {};
     \node [factor, fill=red!100, between=x5t and T5] (ft5) {};
     \node [factor,  left=of T5] (f03) {};
     \node [factor, fill=blue!100, between=x3t and x5t] (fs35) {};

     \node [latent, left=of x3t, yshift=0.6cm, xshift=-0.9cm] (x3cf) {$x_3^i$};

     \node [latent, right=of x3cf, xshift=-0.25cm] (x5cf) {$x_5^i$};
     
     \node [factor, fill=blue!100, between=x3cf and x5cf] (fs35cf) {};
     \node [factor, fill=orange!100, between=x3cf and x5cf, yshift=-0.5cm] (ft35cf) {};
     \node (e2)[const] at (17.35, 1.6)  {};

      \node (x3)[latent] at (15.1,-1.0) {$x_3^i$};
      \node [factor, right=of x3] (f1) {};
      \node [latent, right=of f1, xshift=-0.55cm] (x4) {$x_4^i$};
      \node [factor, right=of x4] (f2) {};
      \node [latent, right=of f2, xshift=-0.55cm] (x5) {$x_5^i$};
      \node [factor, fill=orange!100, between=x3 and x5, yshift=0.75cm] (fm35) {};

       \node [draw, fill=black!100,circle,minimum size=4pt, left=of x3,xshift=0.75cm,inner sep=0pt] (dot0) {};
      \node [draw, fill=black!100,circle,minimum size=4pt, left=of x3,xshift=0.5cm,inner sep=0pt] () {};
      \node [draw, fill=black!100,circle,minimum size=4pt, left=of x3,xshift=0.25cm,inner sep=0pt] (dot1) {};

     \node [latent, below=of x4,xshift=-0.0cm] (l1) {$M^i$};
     \node [factor,  between=x3 and l1] (fl1) {};
     \node [factor, between=x4 and l1] (fl2) {};
     \node [factor,  between=x5 and l1] (fl3) {};

     \begin{pgfonlayer}{background}
	\draw[fill=blue!10,rounded corners] (13.6,-3.25) rectangle (19.1,0)
        node[below, xshift=-4cm, yshift=-2.75cm] {SLAM Module};
        \draw[fill=green!10,rounded corners] (13.6,3.25) rectangle (19.1,0.5)  node[above, xshift=-4cm, yshift=2.2cm] {Tracking Module};
        
        \draw[fill=red!10,rounded corners] (16.15,2.5) rectangle (13.75,0.65) node[above, xshift=0.35cm, yshift=1.35cm] {CF};
        \draw[dashed] (18.1, 1.85) circle [x radius=0.9cm, y radius=3mm] ;

     \end{pgfonlayer}

  \end{scope}
  
 \graph {
    (import nodes);
    x3--[dashed]dot0,
    x3t--ft3, ft3--T5, f03--T5, x3t--fs35, x5t--fs35,
    {x5t,T5}--ft5, 
    x3--f1, f1--x4, x4--f2, f2--x5,
    {x3,l1}--fl1, {x4,l1}--fl2,{x5,l1}--fl3,
    x3--[  bend left=20]fm35,
    x5--[  bend right=20]fm35,
    {x3cf,x5cf}--fs35cf,
    x3cf--[ bend right=20]ft35cf,
    x5cf--[ bend left=20]ft35cf,
    ft35cf->[dashed]fm35,
    e2->[dashed]ft35cf,

   };   
\end{tikzpicture}
}

\caption{Robot inner-loop architecture demonstrated with robot $i$. Each robot has a SLAM module (purple), a Tracking module (green), and a CF between the SLAM and tracking modules (pink). (a) and (c): When the robot measures the target, e.g., time steps 3 and 5, it requests and receives a pose estimate from the SLAM module (blue factors). (b) and (d): New data from the tracking module, due to target measurement or communication with neighboring robots, flows back to the SLAM module via the orange factors. }
\label{fig:archit_full}

\end{figure*}
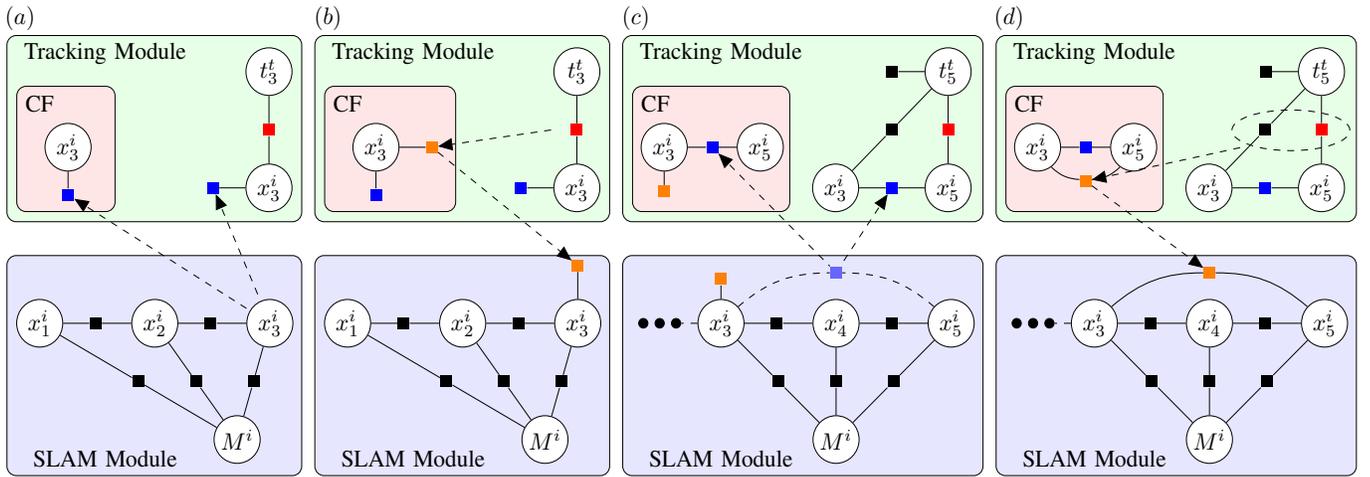

Second, we can again exploit conditional independence, now in the inner-robot level, to separate the map $M^i$ from the targets $T^i$. 
Looking at the two left boxes (purple and green) in Fig. \ref{fig:approachGraph}, the map and tracking variables are conditionally independent given the ego poses in which the targets were observed, i.e. in this example $M^i \perp T^i|x_3^i$.
With this conditional independence, drawing inspiration from the \emph{NET-DDF} architecture suggested in \cite{loefgren_scalable_2020}, we can separate between the SLAM and tracking solutions. 
As shown in Fig. \ref{fig:archit_full}, we suggest using two modules - a tracking module, to handle the robot's tracking assignment, and a SLAM module, to solve for the robot's pose and the local map. 
This architecture contains a filter and a channel filter (CF) \cite{grime_data_1994} to explicitly track the data transferred between the local filter, responsible for tracking, and the SLAM module.
The architecture expresses the following factorization of (\ref{eq:decSLAM}) and data fusion between the two modules,
\begin{equation}
    \begin{split}
    &p^i(\chi^i|Z^i_k)=\\    &\frac{\overbrace{p(M^i|x^i_{k:0},Z^{i,s}_k)p(x^i_{k:0}|Z^{i,s}_k)}^{\textrm{SLAM Module}}\cdot \overbrace{p(x^i_{K}|Z^{i,t}_k)p(T^i_{k}|x^i_{K},Z^{i,t}_k)}^{\textrm{Tracking Module}}}
    {\underbrace{p(x^i_{K}|Z^{i,s}_k\cap Z^{i,t}_k)}_{\textrm{CF}}}.
     \end{split}
    \label{eq:decSLAMfactorized}
\end{equation}
Where $Z_k^{i,s}$ and $Z_k^{i,t}$ are the data sets gathered by the SLAM and tracking modules, respectively, and $K$ describes a vector of time steps in which the targets were observed. 

The pipeline then works in the following way: 
\begin{enumerate}
    \item \textbf{Initialization:} Each robot $i$ instantiates (i) a local SLAM engine (purple in Fig. \ref{fig:archit_full}), (ii) a tracking filter (green in Fig. \ref{fig:archit_full}) to estimate the set of $N_t^i$ targets, represented by a factor graph, (iii) a stack of CFs, one for the SLAM-Tracking data tracking (pink in Fig. \ref{fig:archit_full}) and another one for each communication link with neighboring robots (not shown), to track common data regarding common targets \cite{dagan_factor_2021}, \cite{dagan_conservative_2022}.    
    \item \textbf{Run SLAM:} Each robot $i$ uses its' local SLAM algorithm to maintain a map of the environment and an ego pose estimate (Fig \ref{fig:archit_full}(a)).
    \item \textbf{Request pose estimate:} When the robot observes a target $t\in T^i$, the tracking module `asks' the SLAM module for a pose estimate. 
    \item \textbf{Send pose estimate:} The SLAM module sends a factor (or a set of factors) proportional to the marginal pose estimate, e.g., the blue factor in Fig. \ref{fig:archit_full}(a) and (c). 
    \item \textbf{Receive pose estimate:} After removing common data by the CF, a new factor is integrated into both the CF and the tracking module's graphs.  
    \item \textbf{Peer-to-peer fusion:} The tracking module can now communicate with a neighboring robot over common targets according to Fig. \ref{fig:approachGraph}, and using the following version of the heterogeneous fusion rule (\ref{eq:Heterogeneous_fusion}), 
    \begin{equation}
    \begin{split}
        p_f^i(x^i_{K}, T^i&|Z^{i,t,+}_k)\propto 
        p^i(x^i_{K},T^{i\backslash j}|T^{ij}_{C,k},Z^{i,t,-}_k)\\ \times &\frac{p^i(T^{ij}_{C,k}|Z^{i,t,-}_k)p^j(T^{ij}_{C,k}|Z^{j,t,-}_k)}{p^{ij}_c(T^{ij}_{C,k}|Z^{i,t,-}_k \cap Z^{j,t,-}_k)}.  
    \end{split}
    \label{eq:Heterogeneous_fusion_tracking}
    \end{equation}
    \item \textbf{Send pose back to SLAM:} The tracking module removes common data using the CF and sends the SLAM module marginal ego pose factors (orange factor in \ref{fig:archit_full}(b) and (d)), as it now has new data to send, both from the robot-target measurement and indirectly from fusion with a neighboring robot $j$. 
    \item \textbf{Add factors to SLAM:} SLAM module adds new data from the tracking module to the SLAM graph (orange factor in \ref{fig:archit_full}(b) and (d)).
    \item \textbf{Repeat:} Process repeats recursively. 
\end{enumerate}


\section{Simulation Plan and Experimental Roadmap}
\label{sec:sim}

To test our approach, we will simulate $n_r\geq 2$ robots moving in a joint Gazebo environment \cite{koenig_design_2004}, tracking $n_t\geq 5$ moving targets.
The robots estimate the $2D$ position and velocity of each target $t^t_k=[X^t_k,\dot{X}^t_k,Y^t_k,\dot{Y}_k^t]^T$. 
Targets are assumed to be moving according to the following linear kinematics model, \begin{equation}
\begin{split}
    &t^t_{k+1}=Ft^t_{k}+Gu^t_k+\omega_k, \ \ \omega_k \sim \mathcal{N}(0,0.08\cdot I_{n_x\times n_x}),\\
    &F=\begin{bmatrix}1 & \Delta t &0 &0\\0 &1 &0 &0\\ 0 &0 &1 & \Delta t\\0& 0 &0 &1 \end{bmatrix}, \quad
    G=\begin{bmatrix}\frac{1}{2}\Delta t^2 &0\\\Delta t&0\\0 &\frac{1}{2}\Delta t^2\\0 &\Delta t \end{bmatrix}.
    \end{split}
    \label{eq:dynamicEq}
\end{equation}
Where $\Delta t$ is the time step, $u_k^t$ is the control input of target $t$, and $\omega_k$ is zero-mean additive white Gaussian (AWGN) noise.
When a target $t$ is within robot $i$'s range, it can take a relative position measurement, 
\begin{equation}
    \begin{split}
        y^{i,t}_{k} &= \begin{bmatrix}
            X^i_k-X^t_k \\
            Y^i_k-Y^t_k
            \end{bmatrix} +v^{i}_k, \ \ v^{i}_k \sim \mathcal{N}(0,R^{i}_k),
    \end{split}
    \label{eq:meas_model}
\end{equation}
where $v^i_k$ is zero-mean AWGN noise. For now, it is assumed that the target data association problem is solved separately.  

To test our system, we need a dataset that includes data describing (i) multiple robots in a joint environment, and (ii) different measurement modalities, e.g., LiDAR and camera.  
To the best of our knowledge, such a dataset does not exist. For example, \cite{burri_euroc_2016} contains several sequences that can be used in parallel to simulate collaborative SLAM, but all sequences contain only visual-inertial data. 
On the other hand, \cite{yin_m2dgr_2021} provides data from multiple sensors such as IMU, LiDAR, and Camera, but it only includes one sequence per scenario, which is not applicable for a multi-robot scenario. 
Thus, to rigorously build and test our system, our plan is to start with Gazebo-based simulations before moving to hardware experiments. 
We designed a simple Gazebo environment, including Clearpath Jackal unmanned ground vehicles (UGV) that can simulate the SLAM and tracking robots, as well as the targets.
The preliminary study aims to test the architecture of the system, we will use LIO-SAM \cite{shan_lio-sam_2020}, a factor graph-based lidar-inertial odometry system, as the SLAM engine. 
When the tracking module requests a pose estimate from the SLAM module (see \ref{sec:techApproach}), we query the LIO-SAM factor graph for the marginal distribution over the requested time steps, represented by the information vector and matrix.
These are then integrated as a new factor to the tracking module factor graph and CF, as shown in Fig. \ref{fig:archit_full}(a) and (c). 
When the SLAM module receives a new pose factor back from the tracking algorithm, as shown in Fig. \ref{fig:archit_full}(b) and (d), it adds it to the LIO-SAM graph as a GPS or loop closure type factors \cite{shan_lio-sam_2020}.  

When successful, the rest of the test plan is as follows: (i) Gazebo simulation with two different SLAM algorithms, e.g., LIO-SAM \cite{shan_lio-sam_2020} - LiDAR-inertial, and Kimera \cite{rosinol_kimera_2021} - visual-inertial. (ii) hardware experiments using 2-3 Clearpath Huskey UGVs and varying numbers of maneuvering and static targets in an outdoor environment.






\section{Summary}
\label{sec:summary}
The primary goal of this paper is to report progress toward a heterogeneous robotic system, where robots are able to collaborate despite differences in their onboard algorithms.
We use factor graphs to analyze and exploit the conditional independence structure inherent in the decentralized multi-robot heterogeneous SLAM and tracking problem. 
We are then able to design a system architecture that separates the SLAM and tracking solutions of the robot and between the SLAM systems used by neighboring robots.
While this paper focuses on a SLAM and target tracking problem, this approach can be extended to other robotic applications such as cooperative localization and navigation \cite{dourmashkin_gps-limited_2018,loefgren_scalable_2019}, and terrain height mapping \cite{schoenberg_distributed_2009}.







\IEEEtriggeratref{20}

\bibliographystyle{IEEEtran}
\bibliography{references.bib}

\end{document}